\documentclass[letterpaper]{article}
\usepackage{proceed2e}
\usepackage[margin=1in]{geometry}
\usepackage{subcaption}
\usepackage{amsmath}
\usepackage{amsfonts}
\usepackage[round]{natbib}

\usepackage[]{algorithm2e}
\usepackage{amsthm}
\usepackage{color}
\usepackage{graphicx}
\usepackage{listings}
\usepackage{pgfplots}

\usepackage{mathrsfs}
\usepackage{tikz}
\usetikzlibrary{arrows.meta}
\usetikzlibrary{patterns}
\usetikzlibrary{automata,positioning}
\usepackage[T1]{fontenc}
\usepackage{inconsolata}
\usepackage{sidecap}    
\usepackage{mathtools}

\usepackage{lineno}

\newtheoremstyle{exampstyle}
  {\topsep} 
  {1pt} 
  {} 
  {} 
  {\bfseries} 
  {.} 
  {.5em} 
  {} 
\theoremstyle{remark}

\theoremstyle{exampstyle}
\newtheorem{definition}{Definition}[section]
\newtheorem{theorem}{Theorem}[section]
\newtheorem{proposition}{Proposition}[section]

\usepackage{times}

\title{Probabilistic Program Abstractions}

\author{ 
{\bf Steven Holtzen} \and {\bf Todd Millstein} \and {\bf Guy Van den Broeck} \\
Computer Science Department \\
University of California, Los Angeles\\
\texttt{\small \{sholtzen,todd,guyvdb\}@cs.ucla.edu} 
}

\newcommand{\dbracket}[1]{[\![ #1 ]\!]}
\def\Pr{\mathop{\rm Pr}\nolimits} 
\newcommand{\prob}[0]{\Pr} 
 
\newcommand{\abs}[0]{\mathcal{A}}
\newcommand{\dabs}[0]{\mathcal{D}_\abs}
\newcommand{\conc}[0]{\mathcal{C}}
\newcommand{\concd}[0]{\mathcal{D}_\conc}
\newcommand{\eabs}[0]{\dbracket{\abs}}
\newcommand{\econc}[0]{\dbracket{\conc}}

\newcommand{\pow}[1]{2^{#1}} 

\newcommand\myeq{\stackrel{\mathclap{\scriptsize\normalfont\mbox{def}}}{=}}

\newcommand{\guy}[1]{}
\newcommand{\steven}[1]{}  
\newcommand{\todd}[1]{} 

\begin{document}

\maketitle

\begin{abstract}
Abstraction is a fundamental tool for reasoning about complex systems. Program
abstraction has been utilized to great effect for analyzing deterministic
programs. At the heart of program abstraction is the relationship between a
concrete program, which is difficult to analyze, and an abstract program, which
is more tractable. Program abstractions, however, are typically not probabilistic.
We generalize non-deterministic program abstractions to
probabilistic program abstractions by explicitly quantifying the
non-deterministic choices. Our framework upgrades
key definitions and properties of abstractions to the probabilistic context. We also discuss
preliminary ideas for performing inference on probabilistic abstractions and general probabilistic programs.
\end{abstract}

\section{INTRODUCTION \& MOTIVATION}
Program abstractions are a richly studied method from the programming languages
community for reasoning about intractably complex programs~\citep{Cousot1977}.
An abstraction is typically an over-approximation to a program: any execution
that is possible in the original program is contained within the abstraction.
Over-approximation allows abstractions to be used to prove \emph{program
invariants}: any property of all executions in the abstraction is also true of
all executions in the original program. To achieve this goal while being
more tractable than the concrete program, abstractions work on a simplified
domain. The abstraction selectively models particular aspects of the original
program while utilizing non-determinism to conservatively model the rest.

Non-deterministic abstractions are useful for verifying properties such as
reachability in a concrete program. However, abstractions are decidedly not probabilistic: they are concerned with the possible, not the probable.
Therefore, they fail to support more nuanced queries such as probabilistic reachability, or probabilistic program inference. 
We seek to enhance the program abstraction framework by explicitly quantifying the non-deterministic choices made in the abstraction, turning the program abstraction into a probabilistic model.
That is, our probabilistic abstractions are themselves probabilistic programs, which have been the subject of intense study recently (e.g., \citet{Goodman08,Fierens2013,Wood2014,Carpenter2016}).

The key contribution of this paper is the development of a foundational theory
for probabilistic program abstractions. We define probabilistic
abstractions as a natural generalization of traditional abstractions, using random variables as the abstraction mechanism instead of non-determinism. 
We also formalize the relationship between a probabilistic abstraction and a concrete
program, again generalizing from the non-deterministic setting. This includes semantics in both the concrete and abstract domain, the connection between these semantics, and the notion of a sound probabilistic over-approximation. 

A well-known construction of non-deterministic program abstractions is that of a \emph{predicate abstraction}~\citep{GrafSaidi97,Ball2001}. It induces an
abstraction relative to a given set of Boolean predicates about the program
state. We define \emph{probabilistic predicate abstractions}, which are represented by a simple Bernoulli probabilistic program, as an instance of our framework, and a generalization of classical predicate abstraction.

We conclude with a discussion of ideas for performing inference in probabilistic
predicate abstractions, building on {\em model checking} techniques from the programming languages community and {\em weighted model counting} from the artificial intelligence community. 
We then discuss how probabilistic abstractions could be used to simplify inference in probabilistic concrete programs.

\section{NON-DETERMINISTIC PROGRAM ABSTRACTION}
\label{sec:representation}
In this section we provide the semantics and properties of an \emph{over-approximate non-deterministic}
abstraction and provide an example of a particular class of over-approximations
known as predicate abstractions.

\subsection{SEMANTICS AND PROPERTIES} \label{sec:program-abstraction}

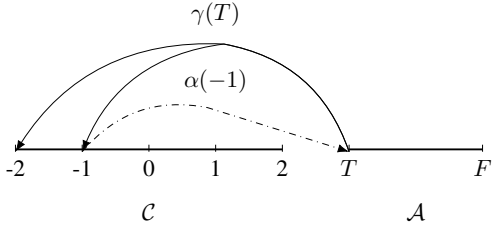
\begin{figure}
\centering
\resizebox{0.85\linewidth}{!}{%
\begin{tikzpicture}
\draw[thick] (0.0,0) -- (4,0);
\draw[xshift=0 cm] (0pt,2pt) -- (0pt,-1pt) node[below,fill=white] (CN2){-2};
\draw[xshift=1 cm] (0pt,2pt) -- (0pt,-1pt) node[below,fill=white] (CN1){-1};
\draw[xshift=2 cm] (0pt,2pt) -- (0pt,-1pt) node[below,fill=white] (C0){0};
\draw[xshift=3 cm] (0pt,2pt) -- (0pt,-1pt) node[below,fill=white] (C1){1};
\draw[xshift=4 cm] (0pt,2pt) -- (0pt,-1pt) node[below,fill=white] (C2){2};

\draw[xshift=2 cm, yshift=-0.7cm] node[below, fill=white]{$\conc$};

\node[label={$\alpha(-1)$}] at (3cm,0.6cm) (P1) {};
\node[label={$\gamma(T)$}] at (3cm,1.6cm) (P2) {};

\draw[thick] (5.0,0) -- (7,0);
\draw[xshift=5 cm] (0pt,2pt) -- (0pt,-1pt) node[below,fill=white] (AN) {$T$};
\draw[xshift=7 cm] (0pt,2pt) -- (0pt,-1pt) node[below,fill=white] (AP) {$F$};

\draw [dash dot,-{Latex}] (CN1.north) to[bend left] (P1) -- (AN.north);
\draw [-{Latex}] (AN.north) to[bend right] (P2) to[bend right] (CN1.north);
\draw [-{Latex}] (AN.north) to[bend right] (P2) to[bend right] (CN2.north);

\draw[xshift=6 cm, yshift=-0.7cm] node[below, fill=white]{$\abs$};

\end{tikzpicture}
}
\caption{Visualization of a simple predicate domain. The five concrete states
over an integer variable $x$ in the range $[-2, 2]$ are abstracted to two states
based on the valuation of the predicate $(x<0)$. We see, for example, that
$\alpha(-1) = T$, and $\gamma(T)= \{-2, -1\})$.
}
\label{fig:nondetabs}
\end{figure}

A concrete program is a syntactic object written $\conc$. The semantics of a
concrete program, which for simplicity we also denote $\conc$, is a function
from input states to output states over some concrete domain $\concd$. Concrete
states are total assignments to all variables in the concrete domain, which we
denote $z \in \concd$.

In general, the problem of proving that a given program satisfies a desired
invariant is undecidable. Advances in theorem proving techniques such as
Satisfiability Modulo Theory (SMT) solvers (e.g., \citet{DeMoura2008}) render
reasoning in many useful theories tractable, yet there exist common program
structures that lie outside of supported theories.

The framework of abstract interpretation~\citep{Cousot1977} provides a general
technique for relating a concrete program $\conc$ to another program $\abs$
which we refer to as an {\em abstraction}. We describe a specialization of the
abstract interpretation framework.

\begin{definition}{\textbf{Abstract semantics of an abstraction.}} The abstract
semantics of an abstraction $\abs$, which for simplicity we also denote $\abs$,
is a function from input states to sets of output states over an abstract domain
$\dabs$, written $\abs : \dabs \rightarrow \pow{\dabs}$.
\label{def:abstract_semantics}
\end{definition}

Intuitively, the nondeterminism in the abstract semantics of an abstraction
represents uncertainty due to the loss of information in abstracting $\conc$ to
$\abs$. We represent this non-determinism as a set of possible abstract states,
denoted $a \in \dabs$. To relate concrete programs with abstractions we
introduce two mappings between concrete and abstract states.

\begin{definition}{\textbf{Abstraction and concretization functions.}} An
\emph{abstraction function} for $\concd$ and $\dabs$ is a function $\alpha :
\concd \rightarrow \dabs$ that maps each concrete state to its abstract
representative. A \emph{concretization function} for $\concd$ and $\dabs$ is a
function $\gamma : \dabs \rightarrow \pow{\concd}$ that maps each abstract state
to a set of concrete states. When applied to sets, $\gamma$ and $\alpha$
respectively concretize or abstract each element of the set.
\end{definition}

Abstraction and concretization functions are related.

\begin{definition}{\textbf{Compatibility.}} An abstraction function $\alpha$ and
concretization function $\gamma$ are {\em compatible} if $z \in
\gamma(\alpha(z))$ for all $z \in \concd$. As an extension, the two functions
are {\em strongly compatible} if they are compatible and for any $a$ and $z \in
\gamma(a)$, we have that $z \notin \gamma(a')$ for any $a' \ne a$.
\end{definition}

A \emph{predicate domain} is a
well-studied abstract domain induced by a given sequence of predicates $(p_1,
\ldots, p_n)$ about the concrete state.
The abstract domain $\dabs$ consists of
$n$ Boolean variables $(b_1, \ldots, b_n)$ and so has $2^n$ possible elements,
one for each valuation to the $n$ variables. 
For instance, suppose $\concd$ consists of a single integer variable $x$ whose
value is in the range $[-2,2]$. The single predicate $(x<0)$ induces an abstract
domain with two possible states, representing the concrete states
where $(x<0)$ is true and false. See Figure~\ref{fig:nondetabs} for a
visualization.
The abstraction function $\alpha$
maps each concrete state $z$ to the abstract state $(p_1(z), \ldots, p_n(z))$,
and the concretization function $\gamma$ maps each abstract state $a$ to the set
of concrete states consistent with it: $\{z \in \concd \mid (p_1(z), \ldots,
p_n(z)) = a\}$. The functions $\alpha$ and $\gamma$ are strongly compatible for predicate domains.
\label{s:predicate_domains}

Intuitively, an abstraction represents a set of possible concrete programs,
which is formalized as follows:
\begin{definition}{\textbf{Concrete semantics of an abstraction.}} The concrete
semantics of an abstraction $\abs$, given compatible abstraction and
concretization functions $\alpha$ and $\gamma$, is a function $\eabs : \concd
\rightarrow \pow{\concd}$ defined as follows:
\begin{align*}
  \eabs(z) = \gamma \big( \abs(\alpha(z)) \big),
\end{align*}
where $\gamma$ is applied to each element of $\abs(\alpha(z))$.
\label{def:concrete-semantics-of-abstraction}
\end{definition}

Ultimately we wish to prove properties about a particular concrete
program~$\conc$ by reasoning about some simpler abstract program $\abs$. From the above
definition of an abstraction's concrete semantics we immediately obtain the
following criterion for relating a specific concrete program $\conc$ to $\abs$:

\begin{definition}{\textbf{Sound over-approximation.}}
Let $\abs$ be some abstract program with compatible abstraction and
concretization functions $\alpha$ and $\gamma$. The tuple ($\abs, \alpha,
\gamma$) is a sound over-approximation of $\conc$ if for all $z \in \concd$,
$\conc(z) \in \eabs(z)$.
\label{def:over_approximation}
\end{definition}

In other words, $\abs$ is sound for $\conc$ if the result of any concrete
execution of $\conc$ is contained within the possible concretizations of the
result of $\abs$ executed on the abstracted input. Sound over-approximations can
be used to verify {\em safety} properties of programs, which intuitively express
the fact that certain ``bad'' things never happen (e.g., no null dereferences
will occur). Every safety property can be formalized as a requirement that some
set $\mathcal{B}$ of ``bad'' states in the concrete program never be reached. To
prove that $\conc(z) \not\in \mathcal{B}$ for each concrete state $z$, it
suffices to prove that $\gamma(\eabs(a)) \cap \mathcal{B} = \emptyset$ for each
abstract state $a \in \dabs$, where $\abs$ is a sound over-approximation of
$\conc$.

In general, the construction of an abstraction is a careful balance between
\emph{precision}, the fidelity of the abstraction to the original concrete program,
and \emph{tractability}, how difficult the abstraction is to construct and
reason about. For abstract predicate domains, adding more predicates to the domain
increases precision but also makes the abstraction more costly to produce and analyze.

The semantics above treats programs $\conc$ and $\abs$ as black-box input-output
functions. Nevertheless, the semantics straightforwardly generalizes to assign meaning to
every single line of code in the programs, allowing us to establish a sound
over-approximation throughout.

\subsection{PREDICATE ABSTRACTION}
\label{sec:predicate_abstraction}

A \emph{predicate abstraction} is a well-studied program abstraction whose
abstract domain is a predicate domain~\citep{GrafSaidi97,Ball2001} (see
the previous section for the definition of a predicate domain).
Predicate abstractions are known as {\em Boolean programs}: the domain $\dabs =
\{T,F\}^n$. Safety checking in Boolean programs is  decidable: a
Boolean program has a finite set of states over a fixed number of Boolean
variables, making it decidable to obtain the set of reachable states. Given a
concrete program $\conc$ and a set of $n$ predicates $(p_1, \ldots, p_n)$ over
the concrete domain~$\concd$, the goal of the predicate abstraction process is
to construct an abstract Boolean program $\abs$ that forms a sound
over-approximation of $\conc$ and is as precise as possible relative to the given predicates.

\begin{SCfigure}[1.2]
  \centering
\begin{lstlisting}[numbers=left,mathescape=true,xleftmargin=5.0ex]
if(x<0) {
  x = 0
} else {
  x = x + 1
}
\end{lstlisting}
~~~~
  \caption{A simple concrete program over an integer variable \texttt{x}.}
  \label{fig:concrete}
\end{SCfigure} 

\begin{figure}
\centering
\begin{lstlisting}[numbers=left,mathescape=true,xleftmargin=3.5ex]
if(*) {
  assume({x<3})
  {x<-4}, {x<3} = $F$, $T$
} else {
  assume(!{x<-4})
  {x<-4}, {x<3} = 
     choose($F$, !{x<3} $\vee$ !{x<-4}), 
     choose({x<-4}, !{x<3})
}
\end{lstlisting}
  \caption{A predicate abstraction of the program in Figure~\ref{fig:concrete}
induced by the predicates {\tt x<-4} and {\tt x<3}. Note that predicate updates that are abstractions of the same concrete assignment statement are
updated simultaneously.}
  \label{fig:abstract}
\end{figure} 

We use the simple program in Figure~\ref{fig:concrete} as an example to illustrate
the predicate abstraction process. The Boolean program induced by the predicates
{\tt x<-4} and {\tt x<3} is shown in Figure~\ref{fig:abstract}.
Following the notation of \citet{Ball2001}, the {\tt *} operator represents
nondeterministic choice, and the Boolean variable associated with predicate
$p$ is denoted {\tt \{$p$\}}. We describe the predicate abstraction process for
branches and assignments in turn.

\subsubsection{Abstracting Branches}
\label{sec:predicate_branch}
Consider a conditional statement of the form 
\begin{center}
{\tt if (p) \{$\cdots$\} else \{$\cdots$\}}
\end{center} 
in the concrete program.
%
Let {\tt p}$^T$ denote the strongest propositional formula over the
predicates $p_1, \ldots, p_n$ that is implied by {\tt p} and {\tt p}$^F$
denote the strongest propositional formula over the predicates $p_1, \ldots,
p_n$ that is implied by {\tt !p}. These formulas represent the most precise
information we can know inside the {\em then} and {\em else} branches
respectively, given the predicates in the abstraction. They can be obtained
through queries to an SMT solver, assuming that {\tt p} and the $n$ predicates
are all in decidable logical theories; see~\citet{Ball2001} for details. The
predicate abstraction process translates the above conditional as follows in the
Boolean program:
\begin{lstlisting}[mathescape=true]
if (*) { 
  assume({p$^T$}) ...
} else { 
  assume({p$^F$}) ...
}
\end{lstlisting}
Here {\tt \{p$^T$\}} is {\tt p$^T$} but with each predicate $p_i$ replaced
by its Boolean counterpart {\tt \{$p_i$\}}, and similarly for {\tt
\{p$^F$\}}.
The statement \texttt{assume}($\varphi$), which is standard in the programming
languages community, silently ignores executions which do not satisfy $\varphi$.
Note that {\tt \{p$^T$\}} and {\tt \{p$^F$\}} can simultaneously be true,
which allows the execution to nondeterministically take either branch of the
conditional.

In the program of Figure~\ref{fig:concrete}, we know that {\tt x<0} is true in the
{\em then} clause. In Figure~\ref{fig:abstract}, the strongest information our
abstraction can know at that point is that (the Boolean variable corresponding
to) {\tt x<3} is true. Similarly, {\tt x<0} is false in the {\em else} branch in
Figure~\ref{fig:concrete}, while the abstraction in Figure~\ref{fig:abstract} only
knows that {\tt x<-4} is false.

\subsubsection{Abstracting Assignment Statements}
\label{sec:predicate_assignment}
Consider an assignment statement of the form {\tt x = e} in the concrete
program. In the corresponding point of the abstract program we must
\emph{simultaneously} update the values of all Boolean variables to reflect the
update to the value of {\tt x}. Suppose we want to update the variable {\tt
\{$p_i$\}}. Let $p_i^T$ denote the weakest propositional formula over the
predicates $p_1, \ldots, p_n$ such that $p_i^T$ holding before the assignment
{\tt x = e} suffices to ensure that $p_i$ will be true after the assignment.
Similarly let $p_i^F$ denote the weakest propositional formula over the
predicates $p_1, \ldots, p_n$ such that $p_i^F$ holding before the assignment
{\tt x = e} suffices to ensure that $p_i$ will be false after the assignment.
Again an SMT solver can be used to obtain these formulas, leveraging the
standard notion of the {\em weakest precondition} of an assignment statement
with respect to a predicate~\citep{dijkstra76}. The predicate abstraction
process updates the Boolean variable {\tt \{$p_i$\}} as follows in the Boolean
program:
\begin{lstlisting}[mathescape=true]
{$p_i$} = choose({$p_i^T$}, {$p_i^F$})
\end{lstlisting}
Here {\tt choose}($\varphi_1$, $\varphi_2$) returns $T$ if $\varphi_1$ is
satisfied, otherwise returns $F$ if $\varphi_2$ is satisfied, and otherwise
chooses nondeterministically between $T$ and $F$.

Consider the assignment statement {\tt x = 0} in Figure~\ref{fig:concrete}. The
abstraction process described above will assign {\tt \{x<3\}} in the Boolean
program to {\tt choose($T$, $F$)}, which simplifies to just $T$ as
shown in Figure~\ref{fig:abstract}. More interestingly, consider the assignment
statement {\tt x = x + 1} in Figure~\ref{fig:concrete}. If {\tt x<-4} is true
before the assignment, then we can be sure that {\tt x<3} is true afterward. If
{\tt x<3} is false before the assignment, then we can be sure that {\tt x<3} is
false afterward. If neither of these is the case, then the abstraction does not
have enough information to know the value of {\tt x<3} after the assignment.
Hence in the Boolean program {\tt \{x<3\}} is assigned to {\tt choose(\{x<-4\},
!\{x<3\})}.

\paragraph{Invariants}
Multiple predicates that involve the same variable are typically constrained in
some way. For example, the predicates \texttt{\{\{x<3\}, \{x<-4\}\}} are
constrained due to the relationship \texttt{\{x<-4\}$\Rightarrow$\{x<3\}}. This
constraint is an invariant which increases the precision of the abstraction
with minimal decrease in tractability. We call this
constraint $\mathcal{I}$, and we can enforce it simply by inserting an
\texttt{assume($\mathcal{I}$)} statement after each set of assignments.

\subsubsection{Proving Program Invariants}
\label{sec:proving_program_invariants}
A predicate abstraction is a sound over-approximation of the original concrete
program. Further, because a Boolean program has a finite set of possible states
at each point in the program, it can be exhaustively explored via a form of {\em
model checking}, which conceptually executes the program in all possible
ways~\citep{Ball2000}. Model checking produces the set of reachable states at
each point in the program, and this information can be used to verify invariants
of the original program.

Consider the Boolean program in Figure~\ref{fig:abstract}. All executions of this
program end in a state where the Boolean variable {\tt \{x<-4\}} has the value
$F$. This implies that {\tt x} always ends in a value greater than or equal
to \texttt{-4} in the original program in Figure~\ref{fig:concrete}. On the other hand, our
predicate abstraction is not precise enough to verify that {\tt x} always ends
in a nonnegative value, though that is true of the original program. A different
choice of predicates would enable such reasoning in the abstraction.

\paragraph{Selecting predicates}
The selection of predicates is clearly a critical component of an effective
predicate abstraction. In this work we focus on the definition and construction of probabilistic predicate abstractions given a fixed set of predicates, leaving automated selection of
predicates for future work.  The programming languages community has developed several
approaches to the problem of predicate selection.  A common approach is to use a form of {\em counterexample-driven refinement}, which iteratively adds predicates until the abstraction is precise enough to prove or disprove the desired property of the concrete program (e.g., \citet{Ball:2002}).  Extending these techniques to the probabilistic
context is a challenging and exciting research problem. 

\section{PROBABILISTIC PROGRAM ABSTRACTION}
The primary contribution of this paper is the extension of the non-deterministic
program abstractions of the previous section to the probabilistic context. We
begin by defining a simple probabilistic programming language. Syntactically,
our probabilistic predicate abstractions will simply be probabilistic programs in this
language. Next, we generalize the abstraction semantics of
Section~\ref{sec:program-abstraction} to the probabilistic context, and define
soundness criteria for probabilistic program abstractions. Finally, we
generalize the
predicate abstraction process from Section~\ref{sec:predicate_abstraction} to
the probabilistic context by placing distributions on the non-deterministic
choices. 

\subsection{PROBABILISTIC PROGRAMMING}
We define a simple
probabilistic programming language, \textsc{Bern}, which contains only (1)
Boolean variables; (2) Boolean operators; (3) Boolean assignments; (4) \texttt{if} statements; (5)
a \texttt{flip}$(\theta)$ operator, which is a Bernoulli random variable with
parameter $\theta$; and (6) an \texttt{observe}($\varphi$) statement, which ignores
executions that do not satisfy some condition $\varphi$. Note that \texttt{observe} statements can
also be captured by a conditional probability query on the distribution.

An extension to \textsc{Bern} is to introduce a \texttt{goto} construct, which would
allow it to reason about underlying concrete programs with arbitrary control
flow. The predicate abstraction framework makes reasoning about loopy concrete
programs tractable~\citep{Ball2001}; however, we defer generalizing the semantics of
loopy probabilistic predicate abstractions to future work.
As an example of a \textsc{Bern} program, one can construct a program that encodes a
Bayesian network $\textcircled{\scriptsize a} \rightarrow \textcircled{\scriptsize b}$:
\begin{lstlisting}[mathescape=true]
a = flip($\theta_1$)
if(a) { b = flip($\theta_2$)} 
else  { b = flip($\theta_3$)}
observe(b)
\end{lstlisting}
This probabilistic program defines the conditional probability of each event by
utilizing the control-flow features of \textsc{Bern}. For example, $\Pr(b \mid
\neg a) = \theta_3$. The \texttt{observe} statement conditions the Bayesian
network on some evidence: thus, queries about $a$ in this program correspond
to $\Pr(a \mid b)$.

Probabilistic programming has proven a natural tool for the construction of
generative statistical models. As such, infrastructure for computing queries on
probabilistic programs has begun to develop in the AI and programming languages communities
\citep{Carpenter2016,Goodman08,Wood2014,Fierens2013}.

\subsection{PROBABILISTIC SEMANTICS}
\label{sec:prob-program-abstraction}
Section~\ref{sec:program-abstraction} identifies both the abstract and concrete
semantics of a program abstraction. We generalize these non-deterministic
semantics to probabilistic semantics by producing families of compatible
probability distributions described by constraints on their support. 

Since syntactically abstractions will be probabilistic programs, the
abstract semantics of a probabilistic abstraction are simply the semantics of
that program, broadly defined.

\begin{definition}{\textbf{Abstract semantics.}} Let $a_i, a_o \in \dabs$. The
abstract semantics of a probabilistic abstraction $\abs$, denoted $\Pr_\abs(a_o
\mid a_i)$, is a \emph{conditional probability distribution} over abstract
domain $\dabs$, which describes the probability of transitioning from an initial
set of states $a_i$ to an output state $a_o$ under the abstraction $\abs$.
\label{def:prob_abstract_semantics}
\end{definition}

To define the concrete semantics of a probabilistic abstraction, we first need
to generalize the concretization function $\gamma$ to the probabilistic context.

\begin{definition}{\textbf{Concretization distribution.}} Let $z \in \concd$ and
$a \in \dabs$. A concretization distribution is a \emph{conditional probability
distribution} $\Pr_\gamma(z \mid a)$ that describes the probability of
concretizing an abstract state $a$ to some concrete state $z$.
\end{definition}

In the non-deterministic setting, we were concerned only with membership in the
set $\gamma$. Here, we generalized $\gamma$ to the probabilistic context by
placing a distribution over possible concretizations.\footnote{For continuous
concrete domains, concretization distributions directly generalize to
concretization densities.} Concretization distributions and abstraction
functions are related as follows:
\begin{definition}{\textbf{Compatibility.}}
An abstraction function $\alpha$ and concretization distribution $\Pr_\gamma$
are \emph{compatible} when, for all $z \in \concd$, $\prob_\gamma(z \mid
\alpha(z)) > 0$. Furthermore, these functions are \emph{strongly compatible} if they are
compatible and for any $a$ and $z$ such that $\Pr_\gamma(z \mid a) > 0$, we have
that $\Pr_\gamma(z \mid a') = 0$ for all $a' \ne a$.
\label{def:prob_compat}
\end{definition}

We are now in a position to define the concrete semantics of a probabilistic
abstraction.

\begin{definition}{\textbf{Concrete semantics.}} Let $z_i, z_o \in \concd$ be
some input and output concrete states. The concrete semantics of an abstraction
$\abs$ given a compatible abstraction function $\alpha$ and concretization
distribution $\Pr_\gamma$ is a \emph{conditional probability distribution}
describing the probability of transitioning from $z_i$ to $z_o$:
\begin{align*}
\Pr_{\eabs}(z_o \mid z_i) = \sum_{a_o \in \dabs} \Pr_\gamma(z_o \mid a_o) \Pr_\abs(a_o \mid \alpha(z_i)). 
\end{align*}\label{def:concrete-semantics-of-prob-abstraction}
\end{definition}

In the case when $\alpha$ and $\Pr_\gamma$ are strongly compatible, we can
refine the above definition:
\begin{proposition}{} Let $z_o,z_i \in \concd$. For strongly compatible $\alpha$
and $\Pr_\gamma$, there exists a single $a_o$ for which $\Pr_\gamma(z_o
\mid a_o) > 0$. Thus the sum may be collapsed:
\begin{align*}
\Pr_{\eabs}(z_o \mid z_i) = \Pr_\gamma(z_o \mid a_o) \Pr_\abs(a_o \mid \alpha(z_i)). 
\end{align*}
\end{proposition} 

As an example, we saw previously that predicate domains allow for strongly
compatible concretization and abstraction functions. We see in
Figure~\ref{fig:probabs} a probabilistic extension to non-deterministic
predicate abstraction. 

Under the probabilistic semantics, we can define a probabilistic analog of
the over-approximation property of $\mathcal{A}$ as a constraint on the support
of $\Pr_{\eabs}$.
\begin{definition}{\textbf{Sound probabilistic over-approximation.}}
Let $\abs$ be a probabilistic program abstraction with compatible abstraction function
$\alpha$ and concretization distribution $\Pr_\gamma$. 
Then the tuple ($\abs$, $\alpha$, $\Pr_\gamma$) is a sound probabilistic
over-approximation of concrete program $\conc$ if for all $z \in \concd$,
$ \Pr_{\eabs}(\conc(z) \mid z) > 0$.
\label{def:prob_over_approximation}
\end{definition}

\subsubsection{Non-Deterministic Semantics}
A sound probabilistic over-approximation is a generalization of a sound
non-deterministic over-approximation in the sense that it provides a
distribution over feasible states. Thus
a sound probabilistic over-approximation has a corresponding sound
non-deterministic over-approximation, which we make precise in the following definitions:

\begin{definition}{\textbf{Non-deterministic semantics.}}
Let $\abs$ be a probabilistic program abstraction with compatible concretization
distribution $\Pr_\gamma$ and abstraction function~$\alpha$. Then there is a corresonding
non-deterministic concretization function $\gamma(a)_\downarrow = \{z \mid
\Pr_\gamma(z \mid a) > 0)\}$ and abstract non-deterministic program
$\abs(a)_\downarrow = \{ a' \mid \Pr_\abs(a' \mid a) > 0)\}$.
\label{def:lowering}
\end{definition}

We observe that $\gamma(a)_\downarrow$ is compatible with $\alpha$ if
$\Pr_\gamma$ is compatible with $\alpha$. Further, soundness of a probabilistic abstraction implies soundness of its corresponding non-deterministic abstraction, and vice versa:

\begin{theorem}{\textbf{Non-deterministic sound over-approximation.}}
For any probabilistic program abstraction $\abs$ with compatible concretization
distribution $\Pr_\gamma$ and abstraction function $\alpha$, the tuple $(\abs,
\alpha, \Pr_\gamma)$ is a sound probabilistic over-approximation to concrete
program $\conc$ if and only if the tuple $(\abs(\cdot)_\downarrow, \alpha,
\gamma(\cdot)_\downarrow)$ is a sound non-deterministic over-approximation to
$\conc$.
\label{thm:over_approx}
\end{theorem}

\subsubsection{Concretization Invariance}
\begin{figure}
  \centering
  \resizebox{0.85\linewidth}{!}{%
\begin{tikzpicture}
\draw[thick] (0.0,0) -- (4,0);
\draw[xshift=0 cm] (0pt,2pt) -- (0pt,-1pt) node[below,fill=white] (CN2){-2};
\draw[xshift=1 cm] (0pt,2pt) -- (0pt,-1pt) node[below,fill=white] (CN1){-1};
\draw[xshift=2 cm] (0pt,2pt) -- (0pt,-1pt) node[below,fill=white] (C0){0};
\draw[xshift=3 cm] (0pt,2pt) -- (0pt,-1pt) node[below,fill=white] (C1){1};
\draw[xshift=4 cm] (0pt,2pt) -- (0pt,-1pt) node[below,fill=white] (C2){2};

\node[label={$\Pr_{\gamma}^1$}] at (3cm,2.2cm) () {};
\node[] at(3cm,2.4cm) (P2) {};
\node[label={$\Pr_{\gamma}^2$}] at (3cm,1.4cm) (P3) {};

\draw[-{Latex},thick,xshift=0 cm] (0pt,0pt) -- (0pt,2cm) node[left] {};
\draw[xshift=0 cm, yshift=2cm] node[above]{$\Pr_{\eabs}$};

\draw[thick, pattern=north west lines] (0,0) rectangle node[above] (TopC1) {} (0.3,0.4);
\draw[thick, pattern=north west lines] (1,0) rectangle node[above] (TopC2) {} (1.3,0.5);

\draw[thick, pattern=north west lines] (0.35,0) rectangle node[above] (TopC3) {} (0.65,0.3);
\draw[thick, pattern=north west lines] (1.35,0) rectangle node[above] (TopC4) {} (1.65,0.7);

\draw[thick, fill={rgb:black,1;white,3}] (2,0) rectangle (2.3,1.1);
\draw[thick, fill={rgb:black,1;white,3}] (3,0) rectangle (3.3,1.2);
\draw[thick, fill={rgb:black,1;white,3}] (4,0) rectangle (4.3,0.8);

\draw[xshift=2 cm, yshift=-0.7cm] node[below, fill=white]{$\conc$};
\draw[thick, pattern=north west lines] (5,0) rectangle node[above] (TopABar) {} (5.3,0.5) ;
\draw[thick , fill={rgb:black,1;white,3}] (6.7,0) rectangle (7,1.9);

\draw[thick] (5.0,0) -- (7,0);

\draw[xshift=5 cm] (0pt,2pt) -- (0pt,-1pt) node[below,fill=white] (AN) {$T$};
\draw[xshift=7 cm] (0pt,2pt) -- (0pt,-1pt) node[below,fill=white] (AP) {$F$};

\draw [dotted,-{Latex}] (TopABar.north) to[bend right] (P2) to[bend right] (TopC1.north);
\draw [dotted,-{Latex}] (TopABar.north) to[bend right] (P2) to[bend right] (TopC2.north);

\draw [dashed,-{Latex}] (TopABar.north) to[bend right] (P3) to[bend right] (TopC3.north);
\draw [dashed,-{Latex}] (TopABar.north) to[bend right] (P3) to[bend right] (TopC4.north);

\draw[-{Latex},thick,xshift=5 cm] (0pt,0pt) -- (0pt,2cm) node[left] {};
\draw[xshift=5 cm, yshift=2cm] node[above]{};
\draw[xshift=5 cm, yshift=2cm] node[above]{$\Pr_{\abs}$};

\draw[xshift=6 cm, yshift=-0.7cm] node[below, fill=white]{$\abs$};

\end{tikzpicture}
}
\caption{Probabilistic predicate abstraction over domain
$\dabs=$\texttt{\{\{x<0\}\}}. Distribution $\Pr_{\eabs}$ over $\concd$ is
generated by (1) a distribution over abstract states $\Pr_\abs$ and (2) one of two
concretization distributions: $\Pr_\gamma^1$ or~$\Pr_\gamma^2$.}
\label{fig:probabs}
\end{figure}

The concrete semantics $\Pr_{\eabs}$ are necessary for reasoning about the
concrete domain. However, directly analyzing $\Pr_{\eabs}$ is made difficult by
the necessity of selecting some compatible concretization distribution
$\Pr_\gamma$. Significantly, in the case when a concrete query can be precisely
represented using a set of abstract states, $\abs$ alone provides sufficient
structure to compute a probability in $\Pr_{\eabs}$ independent of the choice of
$\Pr_\gamma$:
\begin{theorem}{\textbf{Concretization distribution invariance.}} 
Let $\abs$ be a probabilistic program abstraction with strongly compatible concretization
distribution $\Pr_\gamma$ and abstraction function~$\alpha$.
For any $z_i \in \concd$ and $a_o \in \dabs$,
\begin{align*}
  \Pr_{\eabs}(a_o | z_i) \myeq \!\!\!\! \sum_{z_o \in \gamma(a_o)_\downarrow} \Pr_{\eabs}(z_o | z_i) = \prob_\abs(a_o | \alpha(z_i)).
\end{align*} \label{thm:invariance}
\end{theorem}
In other words, the probability of an abstracted event occurring in the concrete
semantics is equivalent to the probability of that event in the abstract
semantics, regardless of the concretization distribution.

We see a visualization of this theorem in
Figure~\ref{fig:probabs}. Regardless of whether $\Pr^1_\gamma$ or $\Pr^2_\gamma$
are chosen,
\begin{align*}
  \Pr_{\eabs} \big(\gamma(\alpha(x=-1))_\downarrow \big) \big) &= \Pr_{\eabs}(\{-1, -2\}) \\
  &= \Pr_{\abs}(\{x<0\}).
\end{align*}

As a consequence, queries performed on the abstraction $\abs$ represent queries
performed on the set of all possible strongly-compatible concretization
distributions. Thus, even though in the probabilistic setting we must reason about a
distribution over concrete states, we can still lift our analyses to the
abstract domain, similar to the benefits of non-deterministic abstraction in
Section~\ref{sec:proving_program_invariants}.

\subsection{PROBABILISTIC PREDICATE ABSTRACTIONS}
Thus far we have seen a semantics for a probabilistic program abstraction, but
we do not yet have a way to generate one for a particular program. In this
section, we seek to generalize predicate abstraction to the probabilistic
domain, and show that in general a probabilistic predicate abstraction is a
family of Boolean probabilistic programs with Bernoulli \texttt{flip}
parameters.

\subsubsection{Branch Statements}
We saw in Section~\ref{sec:predicate_branch} that a predicate abstraction of an
\texttt{if} statement is of the form
\begin{lstlisting}[mathescape=true]
if(*) {assume($\alpha$) $\ldots$ } else {assume($\beta$) $\ldots$}
\end{lstlisting}
where $\alpha$ and
$\beta$ represent the most precise information we can know about the
state of predicates at the \textit{then} and \textit{else} branches of the
program. The behavior of the abstraction is non-deterministic in the case when
both $\alpha$ and $\beta$ hold. A probabilistic predicate
abstraction of this statement should explicitly quantify the probability of choosing
a particular path when either path is possible in the abstraction.

To do so, we first rewrite the predicate abstraction's {\tt if} statement equivalently as follows:
\begin{lstlisting}[mathescape=true]
if$(\neg \beta \lor (\alpha \land *))$ { $\ldots$ } else { $\ldots$ }
\end{lstlisting}
As in the original formulation, this version ensures that the {\em then} clause
will not be taken if $\alpha$ is false and the {\em else} clause will not be
taken if $\beta$ is false.\footnote{Note that by construction $\alpha$ and
$\beta$ cannot both be false.} The non-deterministic choice \texttt{*} then determines
which path to take when both predicates are true.

A probabilistic predicate abstraction must represent a distribution over
paths when $\alpha$ and $\beta$ both hold. Under the semantics of \textsc{Bern}, we may
do so simply by replacing the non-deterministic choice with a {\tt flip}:
\begin{lstlisting}[mathescape=true]
if($\neg\beta \lor (\alpha \, \land \, $flip$(\theta)$)) { ... } else { ... }
\end{lstlisting}
Thus a probabilistic version of the predicate abstraction in Figure~\ref{fig:abstract} would have an {\tt if} statement with guard
\texttt{\{x<-4\}$\lor$(\{x<3\}$\land$flip($\theta$))}, where
$\theta$ represents the conditional probability that the branch is taken given
\texttt{-4 $\le$ x < 3}. As long as $0 < \theta < 1$, all concrete executions
are contained within the support of this probabilistic program abstraction,
implying that it is a sound probabilistic over-approximation.

\subsubsection{Assignment Statements}
\label{sec:indep_prob_assignment}
Section~\ref{sec:predicate_assignment} showed that a concrete assignment is
abstracted to a set of predicate assignments of the form \texttt{$\gamma$ =
choose($\alpha$, $\beta$)}, where $\gamma$ is a predicate and $\alpha$
and $\beta$ encode the most precise update we can make to $\gamma$. The
abstraction behaves non-deterministically: it may assign $\gamma$ to
either \texttt{true} or \texttt{false} when neither $\alpha$ nor $\beta$ holds.
Thus, the probabilistic generalization of an assignment statement needs to
represent the conditional probability of $\gamma$ given $\neg \alpha \land
\neg \beta$.

First, we re-write the \texttt{choose} statement, introducing a
non-deterministic \texttt{*} operator similar to the previous section. We
may write an equivalent update to $\gamma$:
\begin{lstlisting}[mathescape=true]
$\gamma$ = $\alpha \lor (\neg \beta \,\land\, $*$)$
\end{lstlisting}
As above, in  \textsc{Bern} we then replace
\texttt{*} with a Bernoulli random variable:
\begin{lstlisting}[mathescape=true]
$\gamma$ = $\alpha \lor (\neg \beta \,\land\, $flip$(\theta))$
\end{lstlisting}
For example, under this strategy the assignment
statement \texttt{x=x+1} from Figure~\ref{fig:abstract} would be abstracted to the following 
\textsc{Bern} program statements, given predicates \texttt{\{x<-3\}} and
\texttt{\{x<4\}} :
\begin{lstlisting}[mathescape=true]
{x<-4}, {x<3} = 
   ({x<-4} $\land$ {x<3} $\land$ flip($\theta_1$)),
   ({x<-4} $\lor$ ({x<3} $\land$ flip($\theta_2$)))
\end{lstlisting}

\subsection{INVARIANTS}
\label{sec:predicate_constraints}
In the non-deterministic case, enforcing invariants among predicates is a
lightweight procedure of inserting \texttt{assume} statements in order to
increase the precision of the abstraction. Analogously, in the probabilistic
case, we wish to represent distributions over predicates while disallowing
inconsistent predicate states. In this section we explore the consequences of
enforcing invariants on the abstraction.

An initial approach to enforcing invariants is to straightforwardly generalize
the non-deterministic procedure by inserting \texttt{observe($\mathcal{I})$}
statements between each assignment, where $\mathcal{I}$ is the invariant which
must hold over the predicates. For example, for the concrete program
\texttt{x=x+10} with the predicates \texttt{\{x<-4\}} and \texttt{\{x<3\}}, we
generate the following abstraction:
\begin{lstlisting}[mathescape=true]
{x<-4}, {x<3} = 
    ({x<-4} $\land$ {x<3} $\land$ flip($\theta_1$)),
    ({x<-4} $\land$ {x<3} $\land$ flip($\theta_2$))
observe({x<-4}$\Rightarrow${x<3})
\end{lstlisting}

A key downside is that the parameters no longer have a
\emph{local semantics}: conditioning correlates the otherwise independent flips.
This complicates the probability computation, which now involves a partition function.

Therefore we present an alternative abstraction construction procedure which
preserves the local semantics of the parameters of the abstraction while
enforcing invariants over predicates. Consider
again the concrete program \texttt{x=x+10}. We generate an abstraction using the
same predicates as before. However, instead of simply inserting \texttt{observe}
statements, we utilize control flow in order to effectively condition on the
previously assigned value:
\begin{lstlisting}[mathescape=true]
{x<3} = {x<3} $\land$ {x<-4} $\land$ flip$(\theta_1)$
if({x<3}) {
  {x<-4} = {x<-4} $\land$ flip$(\theta_2)$
} else {
  {x<-4} = $F$
}
\end{lstlisting}
This abstraction, which we call \emph{structurally dependent}, updates each
predicate sequentially, considering all previous decisions. Each concrete
statement is abstracted to several abstract statements which utilize
control flow to disallow invalid states. The state
\texttt{\{x<-4\}$\land$!\{x<3\}} is guaranteed to have 0 probability without the
use of \texttt{observe} statements. Further, the parameters have a local
interpretation as a conditional probability: it is not necessary to compute a partition function to compute
the probability of a particular predicate configuration.

Fundamentally, these two methods of constructing the abstraction represent
different factorizations of the distribution. In the non-deterministic context
with invariant enforcement,
these two abstractions are equivalent.

\section{DISCUSSION}

This paper focuses on the definition and key properties of probabilistic program
abstractions. In this section we discuss natural next steps for the work.
Traditional non-deterministic program abstractions are typically used to produce
the set of reachable program states, in order to verify invariants. The
analogous operation on a probabilistic program abstraction is inference. First
we discuss possible approaches to inference for probabilistic predicate
abstractions, by leveraging both model checking and weighted model counting.
Second, we discuss how the ability to perform inference on a probabilistic abstraction
could be a key enabler for a new approach to performing inference on more general
probabilistic programs. The main idea is to reduce inference on a probabilistic
program to the task of choosing particular {\tt flip} probabilities for a corresponding
probabilistic abstraction.


\subsection{INFERENCE FOR PROBABILISTIC PREDICATE ABSTRACTIONS}

We believe that existing techniques from the programming languages literature which are
designed for working with non-deterministic Boolean programs can be extended to perform inference on
\textsc{Bern} programs. We can then use weighted model counting to evaluate queries. We note
that abstractions allow one to query the marginal probability of an event at any
point in the program, not merely upon program termination.

\paragraph{Probabilistic Model Checking}
The problem of computing the set of reachable states in a Boolean program is
known as the \textit{model checking problem} and has been extensively studied
by the programming languages community. Commonly one represents
the set of reachable states at any point in the program as some Boolean
knowledge base $\Delta$. In many existing tools, $\Delta$ is represented using a
binary decision diagram~\citep{Ball2000}. Inference in \textsc{Bern} is thus an extension
to the traditional model checking paradigm in which we introduce weighted
variables for the state of each \texttt{flip}. During model checking, we treat
each \texttt{flip} as an unconstrained Boolean variable.

For example, consider the probabilistic predicate abstraction statement
\texttt{\{x<4\} = \{x<4\} $\land$ flip($\theta$)}. We assume \texttt{$\Delta$ =
\{x<4\}} prior to execution of statement. Following this statement,
\texttt{$\Delta'$ = (\{x<4\}$\land$flip($\theta$)) $\lor$ (!\{x<4\} $\land$
  !flip($\theta$))}. See \cite{Ball2000} for more details. 

\paragraph{Weighted Model Counting}
Whereas model checking is usually concerned with determining whether $\abs$ can
reach a particular state, in probabilistic program inference we are concerned
with the weighted sum of reachable states, where the weights are induced by the
parameters of the \texttt{flip}s in each model. The programming languages
community has two primary methodologies for computing the set of reachable
states in a Boolean program: (1) knowledge compilation to binary decision
diagrams~\citep{Ball2000}, and (2) satisfiability methods~\citep{Donaldson2011}.
Both of these approaches can be generalized to perform weighted model counting
for inference in \textsc{Bern}.

The knowledge compilation approach to model checking is already capable of
performing weighted model counting due to the nature of the queries efficiently
supported by a binary decision diagram~\citep{Darwiche2001}, and is used for
inference in discrete probabilistic programs~\citep{Fierens2013} and Bayesian
networks~\citep{Chavira2008}. The satisfiability approach to model checking can
be extended to perform weighted model counting. This problem is \#P-hard~\citep{Valiant1979}, but a number of recent
approximation methods have been explored~\citep{Chakraborty2013,BelleUAI15,Zhao2016}; see
\citet{GomesSS09} for a survey of the subject.

\begin{figure}
    \begin{subfigure}[b]{0.5\textwidth}
\begin{lstlisting}[mathescape=true, numbers=left, xleftmargin=5.0ex]
a = unif [0, 10)
if (a < 5) { b = unif [0, 10) } 
  else { b = unif [0, 20) }
if (b < 5) { c = unif [0, 10) } 
  else { c = unif [0, 20) }
\end{lstlisting}
      \caption{Probabilistic program for Bayesian network
        $\textcircled{\scriptsize a} \rightarrow \textcircled{\scriptsize b} \rightarrow \textcircled{\scriptsize c}$.}
      \label{fig:concrete_prob}
    \end{subfigure}
      \begin{subfigure}[b]{0.5\textwidth}
\begin{lstlisting}[mathescape=true]
{a<5} = flip(1/2)
if({a<5}) { {b<5} = flip(1/2) }
  else { {b<5} = flip(1/4) }
if({b<5}) { {c<5} = flip(1/2) }
  else { {c<5} = flip(1/4) }
\end{lstlisting}
        \caption{Probabilistic abstraction with \texttt{\{a<5\}}, \texttt{\{b<5\}}, and \texttt{\{c<5\}}.}
      \label{fig:interesting_abstraction}
      \end{subfigure}
      \caption{A concrete probabilistic program and a probabilistic abstraction for computing $\Pr_{\eabs}(c<5)$.}
    \end{figure}

\subsection{INFERENCE FOR GENERAL PROBABILISTIC PROGRAMS}
Consider the probabilistic program in Figure~\ref{fig:concrete_prob} and suppose
we want to evaluate $\Pr_{\econc}(c<5)$. We will sketch an approach to doing so
using probabilistic predicate abstractions.

Figure~\ref{fig:interesting_abstraction} shows a probabilistic predicate
abstraction for our original probabilistic program, induced by the predicates
\texttt{\{a<5\}}, \texttt{\{b<5\}}, and \texttt{\{c<5\}}. Initially each {\tt
flip} has its own parameter to represent its probability. In the figure, we show
particular values for each parameter, which were computed by performing queries
on fragments of the original concrete program. For example, the concrete
assignment \texttt{a = unif[0, 10)} is abstracted to \texttt{\{a<5\} =
flip(1/2)} by computing $\Pr(a<5)$ on this single statement of the concrete
program. The other parameters can be learned similarly. The key point is that
each of these queries is much easier to evaluate in the original program than
the actual query of interest, as they are over smaller fragments of the program.

Now we show that the abstraction captures enough detail to answer our query
precisely. Computing the weighted model count using the approach described in
the previous subsection, we see that:
\begin{align*}
  & \Pr_{\abs}(\texttt{\{c<5\}}) =
  \overbrace{0.5\cdot0.5\cdot0.5}^{\texttt{\{a<5\},\{b<5\},\{c<5\}}} +
  \overbrace{0.5\cdot0.5\cdot0.25}^{\texttt{\{a<5\},!\{b<5\},\{c<5\}}} \\
 & \quad +  \underbrace{0.5\cdot0.25\cdot0.5}_{\texttt{!\{a<5\},\{b<5\},\{c<5\}}} +
   \underbrace{0.5\cdot0.75\cdot0.25}_{\texttt{!\{a<5\},!\{b<5\},\{c<5\}}} = \frac{11}{32}.
\end{align*} 
The result is in fact the answer to the original query. 

In this way, the inference problem on $\conc$ is decomposed into two,
potentially much simpler, problems: (i) fixing the parameters of an abstraction,
and (ii) weighted model counting on the abstraction. There remains considerable
theoretical work to formally connect the semantics of the probabilistic
abstraction with a probabilistic concrete program, as well as practical work to
realize the benefits of the approach on desired applications.

\section{RELATED WORK}

\paragraph{Probabilistic reasoning and static analysis.} Several recent works
leverage a probabilistic model to guide refinements of a program
abstraction~\citep{Grigore2016,Zhang:2017}. However, the abstractions themselves
are not probabilistic. \citet{Gehr2016} use static analysis of a
probabilistic program to decompose the problem of inference along paths, which
are then dispatched to specialized integration tools depending on the
constraints of each path; this work analyzes the original concrete program and
does not rely on abstractions.

\paragraph{Probabilistic abstract interpretation.} Probabilistic abstract
interpretation is used to reason about programs with probabilistic semantics,
for example to place upper bounds on the probability of a particular
path~\citep{Monniaux2000} or construct Monte-Carlo methods~\citep{Monniaux2001};
this line of work does not explore the connections between abstractions and
probabilistic programs, nor does it model concrete program marginals. However,
our work does not reason about unbounded loops. The framework of
\citet{Cousot2012} is a highly general framework for reasoning about programs
using probabilistic abstract interpretation; however, they do not consider the
abstraction itself to be a statistical model.

\paragraph{Probabilistic programming systems.} Many systems have been developed
within the AI and programming languages communities that tackle the problem of
probabilistic program inference, but few utilize abstractions. Systems such as
Church \citep{Goodman08}, Anglican \citep{Wood2014}, Stan \citep{Carpenter2016},
BLOG \citep{Milch05}, and others directly analyze the concrete program. Weighted
model counting and knowledge compilation have been used to perform probabilistic
program inference~\citep{Fierens2013}; they also do not leverage program
abstractions. Several probabilistic inference approaches capture distributions
in continuous domains by using Boolean predicates, either as an
approximation~\citep{Michels2016} or as an exact
representation~\citep{BelleIJCAI15}. Finally, program abstraction with the
purpose of inference is an instance of approximate lifted
inference~\citep{kersting2012lifted}: the abstract domain groups together sets
of concrete states, with the aim of reasoning at the higher level.



%
%

\section{CONCLUSION}
Probabilistic program abstractions are currently unexplored territory for aiding
in the analysis of programs, despite the popularity of probabilistic programming. 
We provided a formal framework, derived useful properties, and described probabilistic predicate abstractions techniques.
Much theoretical and practical work remains to be done in exploring
alternative characterizations,
showing relationships between concrete programs and their abstractions, and building practical probabilistic abstraction tools. We hope our framework provides the foundational theory to enable these advances in the future.

\subsubsection*{Acknowledgements}

This work is partially supported by NSF grants \#CCF-1527923, \#IIS-1657613, and
\#IIS-1633857, and by DARPA grant \#N66001-17-2-4032. S.H. is
supported by a National Physical Sciences Consortium Fellowship.

\subsubsection*{References}


\bibliographystyle{abbrvnat}
\bibliography{bib}



\end{document}